\documentclass[10pt,twocolumn,letterpaper]{article}

\usepackage{cvpr}              % To produce the CAMERA-READY version
\definecolor{cvprblue}{rgb}{0.21,0.49,0.74}
\usepackage[pagebackref,breaklinks,colorlinks,allcolors=cvprblue]{hyperref}
\usepackage{times}  % DO NOT CHANGE THIS
\usepackage{helvet}  % DO NOT CHANGE THIS
\usepackage{courier}  % DO NOT CHANGE THIS
\usepackage[hyphens]{url}  % DO NOT CHANGE THIS
\usepackage{graphicx} % DO NOT CHANGE THIS
\usepackage{amsmath}
\usepackage{amssymb}
\usepackage{amsthm}
\usepackage{subcaption}
\usepackage{booktabs}
\usepackage{multirow} %tabllular
\usepackage{array}
\usepackage{pifont}% http://ctan.org/pkg/pifont
\usepackage{algorithm}
\usepackage{algorithmic}
\usepackage[accsupp]{axessibility}

\usepackage{xcolor}  
\definecolor{steelblue}{RGB}{70,130,180}
\usepackage{graphicx}
\def\paperID{****} % *** Enter the Paper ID here
\def\confName{CVPR}
\def\confYear{2026}

\title{FedDAP: Domain-Aware Prototype Learning for Federated Learning under Domain Shift}

\author{Huy Q. Le\textsuperscript{1}, Loc X. Nguyen\textsuperscript{2}, Yu Qiao\textsuperscript{2},  Seong Tae Kim\textsuperscript{2}\footnotemark[1], Eui-Nam Huh\textsuperscript{2}\footnotemark[1], Choong Seon Hong\textsuperscript{2}\thanks{Corresponding author.}\\
\textsuperscript{1}G-LAMP NEXUS Institute, Kyung Hee University \hspace{2mm} \textsuperscript{2}Kyung Hee University  \\
{\tt\small \{quanghuy69, xuanloc088, qiaoyu, st.kim, johnhuh, cshong\}@khu.ac.kr} 
}

\begin{document}
\maketitle
% \renewcommand*{\thefootnote}{\fnsymbol{footnote}}
% \setcounter{footnote}{1}
% \footnotetext{Corresponding Author.}
% \renewcommand*{\thefootnote}{\arabic{footnote}}
% \renewcommand*{\thefootnote}{\fnsymbol{footnote}}

\begin{abstract}
Federated Learning (FL) enables decentralized model training across multiple clients without exposing private data, making it ideal for privacy-sensitive applications. However, in real-world FL scenarios, clients often hold data from distinct domains, leading to severe domain shift and degraded global model performance. To address this, prototype learning has been emerged as a promising solution, which leverages class-wise feature representations. Yet, existing methods face two key limitations: (1) Existing prototype-based FL methods typically construct a \textit{single global prototype} per class by aggregating local prototypes from all clients without preserving domain information.  (2) Current feature-prototype alignment is \textit{domain-agnostic}, forcing clients to align with global prototypes regardless of domain origin. To address these challenges, we propose Federated Domain-Aware Prototypes  (FedDAP) to construct domain-specific global prototypes by aggregating local client prototypes within the same domain using a similarity-weighted fusion mechanism. These global domain-specific prototypes are then used to guide local training by aligning local features with prototypes from the same domain, while encouraging separation from prototypes of different domains. This dual alignment enhances domain-specific learning at the local level and enables the global model to generalize across diverse domains. Finally, we conduct extensive experiments on three different datasets: DomainNet, Office-10, and PACS to demonstrate the effectiveness of our proposed framework to address the domain shift challenges.\footnote{The code is available at \url{https://github.com/quanghuy6997/FedDAP}.} 
\end{abstract}    
\section{Introduction}
\label{sec:intro}

Federated Learning (FL) has been proposed as a decentralized machine learning paradigm that addresses the privacy concerns without sharing private data~\cite{mcmahan2017communication,ye2023heterogeneous}. This privacy-preserving framework is particularly attractive in applications such as healthcare, mobile computing, and sensor networks, where data is inherently distributed~\cite{nguyen2022federated,lee2024federated}. A critical challenge in FL lies in client data's non-independent and identically distributed (non-IID) nature. Data heterogeneity across clients significantly degrades the performance and convergence of the global model~\cite{li2020federated_1,li2022federated,ye2023heterogeneous}.
\begin{figure}[]
	\centering
	\includegraphics[width=\linewidth]{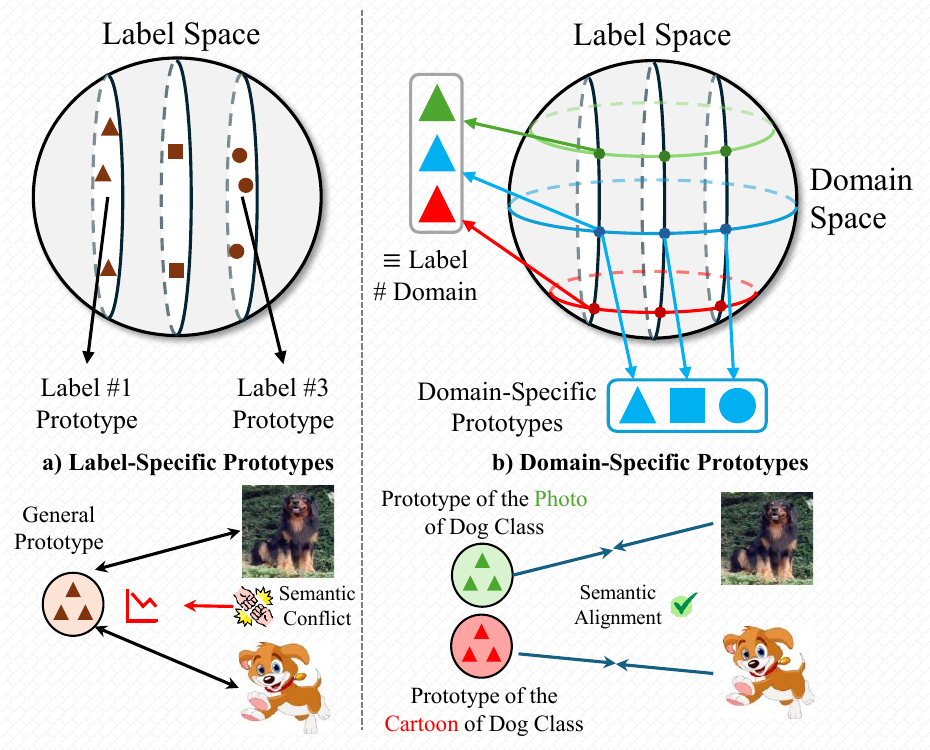}
	\caption{(a) Single label-specific prototypes ignore domain variation, leading to \emph{semantic conflict} when features from different domains are forced to align to general prototype. (b) Our domain-specific prototypes disentangle the feature space along both label semantics (vertical axis) and domain characteristics (horizontal axis) and enabling the better \emph{semantic alignment} to the appropriate prototype. }
	\label{problem_illu} 
\end{figure}
Most existing efforts have addressed this challenge by focusing on label skew, where clients have different class distributions, but often assume that data across clients originates from a shared domain, i.e., the feature distribution for each class remains consistent. However, in real-world scenarios, clients frequently possess data from distinct domains, due to differences in image styles, sensors, environments, or collection modalities, which introduce a domain shift across clients. This \textbf{domain shift} in federated learning leads to varying feature distributions even for the same class, as local data originates from different domains~\cite{gong2022preserving,huang2024federated}.

To address this problem, federated prototype learning~\cite{tan2022fedproto,qiao2023mp,huang2023rethinking,wang2024taming} has recently gained attention as an effective strategy for handling data heterogeneity in FL. Those existing methods enable semantic-level supervision across clients by representing each class using mean feature vectors. This prototype-based approach has been shown to enhance the generalization ability of the global model, especially under non-IID conditions, by providing a compact and class-aware representation that bridges client-specific distributions. While promising, prototype-based FL methods face two fundamental limitations in the presence of domain shift: 
 1) Existing works only construct \textit{a single global prototype per class} by aggregating local prototypes from all clients, without preserving domain information. This approach assumes a domain-agnostic representation per class, which may overlook domain-specific variation. As a result, the global prototype becomes semantically diluted, blending features from diverse domains and may not represent any individual domain accurately. This weakens the effectiveness of local feature alignment and reduces generalization performance. 2) These methods also apply a \textit{domain-agnostic} alignment strategy, forcing all clients to align their features with the same global prototypes, regardless of domain origin. This domain-agnostic supervision ignores the semantic discrepancy between the local client distribution and the global prototype, particularly when they come from different domains. For example, a ``dog" image from the photo domain typically contains natural textures and lighting conditions, while a sketched or cartooned dog may appear in more simplified shapes. Forcing these different representations to align with a single global prototype can result in semantic conflict, which limits the model's generalization across domains.

\begin{figure*}[t]
	\centering
	\includegraphics[width=0.92\linewidth]{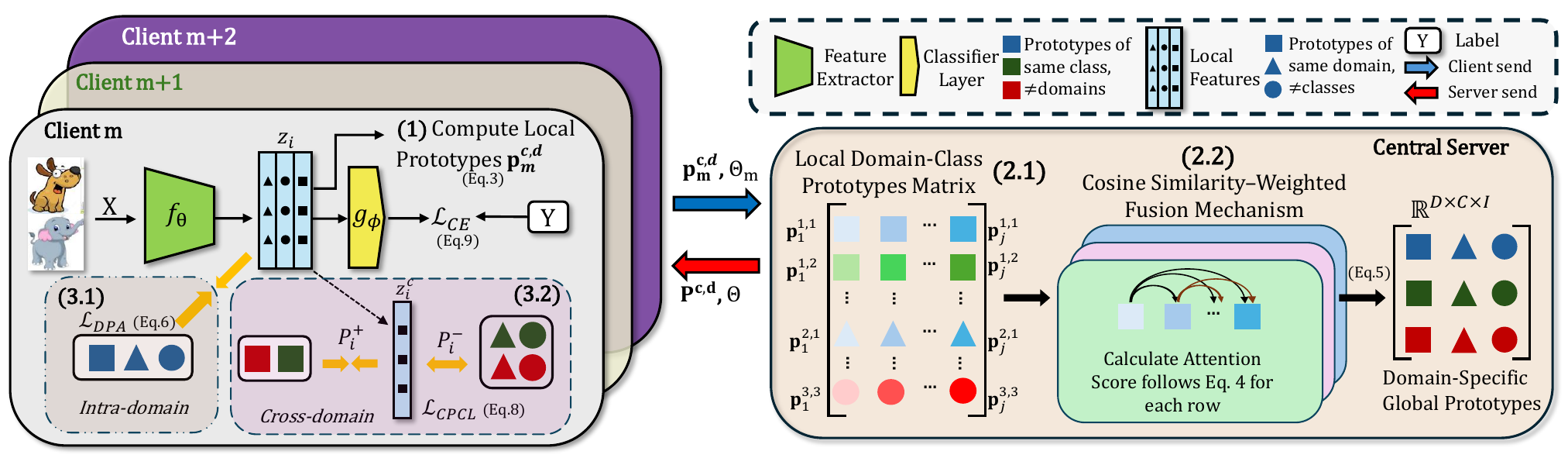}
	\caption{\textbf{Illustration of \textbf{FedDAP}.}  (1) Clients first compute local prototypes $\mathbf{p}_m^{(c, d)}$ based on Eq.~\ref{eq:local_proto} and then upload it to server. (2) The central server performs the cosine similarity-weighted fusion mechanism for the local prototypes by calculating the attention score in Eq.~\ref{eq:attention} for each row of class-domain pair prototypes. Then we perform the Domain-Specific Global Prototype aggregation using Eq.~\ref{eq:domain_proto_agg}. (3) Clients download the global prototypes $\mathbf{P}^{c,d}$ from the server and perform the local training process with the dual prototype alignment strategy with $\mathcal{L}_{DPA}$ in Eq.~\ref{eq:dpa} and $\mathcal{L}_{CPCL}$ in Eq.~\ref{eq:cpcl}. }
	\label{system} 
\end{figure*}

Motivated by these insights, we propose Federated Domain-Aware Prototypes (FedDAP), a novel prototype-based federated learning framework that explicitly incorporates domain knowledge in both prototype construction and alignment to tackle the domain shift problem. As shown in Fig.~\ref{problem_illu}, unlike existing approaches that average prototypes across all clients regardless of domain, we construct a set of domain-specific global prototypes for each (class, domain) pair. These prototypes are generated by aggregating local client prototypes using a cosine similarity-weighted fusion mechanism, which emphasizes semantically consistent representations within same domain and reduces the influence of outlier representations. To fully utilize these prototypes during local training, we introduce a dual prototype alignment strategy. Specifically, the client model first aligns its feature representation with the intra-domain prototype, enforcing semantic consistency and improving the intra-domain stability. Simultaneously, we perform the cross-domain contrastive learning using prototypes from other domains, encouraging the model to generalize across domain boundaries. Under the guidance of dual prototype alignment method, we address the semantic dilution and alignment instability caused by domain-agnostic prototype, achieving the better generalization across heterogeneous clients. Fig.~\ref{system} illustrates the proposed framework of our FedDAP. Our main contributions are:
 \begin{itemize}
     \item We are the first to introduce domain-specific global prototypes, which preserve not only label-space information but also domain-specific semantic representations. This design enables FedDAP to effectively address the domain shift problem in real-world federated learning scenarios.
     \item We propose a novel dual alignment approach that treats intra-domain and cross-domain prototypes differently. Specifically, local features at the client are aligned with intra-domain prototypes to enhance semantic consistency, while cross-domain prototypes are leveraged through contrastive learning to encourage robustness to domain shifts.
     \item  Extensive experiments on three benchmark datasets DomainNet, Office-Caltech, and PACS  demonstrate that our method, FedDAP, consistently outperforms existing prototype-based FL methods, showing improved generalization under domain shift.
 \end{itemize}

%-------------------------------------------------------------------------

\section{Related Work}
\label{sec:formatting}

\subsection{Heterogeneous Federated Learning}
A major challenge in FL is handling non-IID data, where clients have different label distributions or feature characteristics. To address this, several early studies focused on improving local training dynamics through regularization techniques~\cite{karimireddy2020scaffold,li2020federated_1,acar2021federated}, such as proximal objectives, control variates, or adaptive local optimization strategies that aim to reduce divergence between client and global models. On the server side, enhancements in model aggregation~\cite{reddi2021adaptive,Wang2020Federated} have also been proposed, including momentum-based and modular aggregation strategies to stabilize global updates. However, these methodologies generally assume shared data domains among clients. In practice, client data often originates from different domains, introducing domain skew where semantic features of the same class vary significantly across clients due to differences in style and background. To handle such cases, more recent methods have incorporated domain-aware strategies, such as client-specific normalization layers, feature deskewing modules, fairness-aware aggregation, and data augmentation~\cite{li2021fedbn,wang2025federated,zhang2023federated,yan2025simple} to reduce bias introduced by domain divergence. Others have explored domain generalization approaches~\cite{liu2021feddg,le2024efficiently} that aim to learn representations robust to unseen domains, using attention mechanisms or domain-invariant feature modeling.
In contrast to these works, which typically abstract away domain-specific traits or assume generalization to new domains, our method directly models domain-conditioned semantic structure during training. By introducing domain-specific prototypes and a structured dual alignment strategy, our approach is tailored to handle domain shift across training clients, without requiring access to unseen domains or additional auxiliary data.
\subsection{Prototype-driven Federated Learning}
Prototype learning which leverages the class centroids as semantic anchors, has gained traction in tasks like few-shot learning and unsupervised representation learning~\cite{zhang2021prototype,gao2024learning,ma2025protogcd}. In federated learning, this idea was introduced by FedProto~\cite{tan2022fedproto}, enabling clients to communicate lightweight prototypes instead of full models. Recently, FedGMKD~\cite{zhang2024fedgmkd} combined prototype-based knowledge distillation and discrepancy-aware aggregation to boost performance without relying on public datasets. Beyond communication efficiency, recent efforts~\cite{zhang2024fedtgp,zhou2025fedsa} have focused on improving prototype quality under data heterogeneity. Approaches have been developed to reduce prototype redundancy, and enhance semantic consistency, through techniques such as multi-level prototype clustering, adaptive contrastive margins, and sparsity-aware regularization~\cite{tan2022fedproto,tan2022federated,qiao2023mp,wang2024taming}. These works demonstrate that well-designed prototype learning can be both lightweight and robust under non-IID settings. Despite these improvements, existing prototype-driven methods aggregate prototypes without considering domain-specific differences. This domain-agnostic treatment can lead to semantically diluted prototypes and misaligned feature guidance during local training. Moreover, alignment strategies in these methods are typically uniform across clients, ignoring potential mismatch between global prototypes and local domain-specific features. Our method addresses these issues by explicitly constructing domain-specific prototypes, thereby preserving intra-domain semantics. This design improves local stability while leveraging cross-domain information to enhance the global model’s generalization under domain shift.

\section{Methodology}

\subsection{Preliminaries}

We consider a standard FL setup involving a central server and $M$ clients. Each client $m \in \{1, \dots, M\}$ possesses a private local dataset $\mathcal{D}_m = \{(x_i^m, y_i^m)\}_{i=1}^{N_m}$, where $x_i^m \in \mathcal{X}$ {denotes data samples and $y_i^m \in \mathcal{Y}$ represents the corresponding labels. Unlike conventional FL settings that assume data across clients is drawn from the same domain, we consider a more realistic and challenging scenario where client data originates from \textit{multiple domains}. This introduces domain shift across clients, where different clients share a same label space $\mathcal{Y}$ (i.e., $P(y)$ is consistent) while their feature distributions $P(x)$ differ.  Each client model comprises a shared architecture $\Theta$, consisting of a \textit{feature extractor} $f_\theta: \mathcal{X} \rightarrow \mathbb{R}^d$ and a \textit{classifier} $g_\phi: \mathbb{R}^d \rightarrow \mathcal{Y}$, where $\theta$ and $\phi$ denote the model parameters.

Clients perform local training using their private data and periodically communicate updates (e.g., model weights or prototypes) to the server for global aggregation. The global learning objective aims to find a set of parameters $(\theta, \phi)$ that minimizes the weighted sum of local empirical risks across all clients:
\begin{equation}
\min_{\Theta} \sum_{m=1}^{M} \frac{n_m}{n} \cdot \mathcal{L}_m(\Theta),
\end{equation}
where $n_m = |\mathcal{D}_m|$, $n = \sum_{m=1}^{M} n_m$, and the local objective on client $m$ is defined as:
\begin{equation}
\mathcal{L}_m(\Theta) = \frac{1}{n_m} \sum_{(x_i^m, y_i^m) \in \mathcal{D}_m} \mathcal{L}\big(g_\phi(f_\theta(x_i^m)), y_i^m\big),
\end{equation}
with $\mathcal{L}(\cdot, \cdot)$ denoting a task-specific loss function (e.g., cross-entropy). Due to the domain shift nature of the client data, local training objectives diverge significantly, which poses challenges to global optimization and degrades the generalization ability of the aggregated model.

\subsection{Domain-Specific Prototype}\label{proto_agg}
\noindent\textbf{Motivation.}  In prototype-based federated learning, global prototypes are typically constructed by averaging local class prototypes across all clients, implicitly treating class semantics identically across domains. However, when data distributions vary significantly among domains, these global averaged prototypes no longer represent any specific domains, breaking down the above naive approach. To be specific, the data with the same class can be represented by distinct feature patterns across domains due to differences in visual style, modality, or environment. Averaging such diverse representations into a single prototype leads to \textit{semantic dilution}, where the prototype no longer accurately represents any specific domain. This approach oversimplified the representations of prototypes, which weakens its ability to guide the local training while overlooking the domain perspective. To address this limitation, we propose \textbf{domain-specific global prototypes}, where we create a unique prototype for each \textit{class–domain pair} to preserve domain semantic structures during aggregation, instead of maintaining one global prototype per class.

\noindent\textbf{Local Prototype Computation.}
Let each client $m$ have a domain identifier $d \in \{1, \dots, D\}$. For each class $c \in \{1, \dots, C\}$, client $m$ computes a local prototype $\mathbf{p}_m^{(c, d)} \in \mathbb{R}^I$ by averaging the feature embeddings extracted by the shared encoder $f_\theta(x_i^m)\in \mathbb{R}^I$:
\begin{align}
\begin{split}
    \mathbf{p}_m^{(c, d)} &= \frac{1}{|\mathcal{D}_m^{(c)}|} \sum_{(x_i^m, y_i^m) \in \mathcal{D}_m^{(c)}} f_\theta(x_i^m)
    \\
    \mathbf{p}_m&=[\mathbf{p}_m^{(1,d)},\mathbf{p}_m^{(2,d)},\dots,\mathbf{p}_m^{(C,d)}]  \in \mathbb{R}^{C \times I},
\label{eq:local_proto}
\end{split}
\end{align}
where $\mathcal{D}_m^{(c)} \subset \mathcal{D}_m$ denotes the subset of client $m$'s data belonging to class $c$. 

\noindent\textbf{Domain-Specific Global Prototype Aggregation.}
Clients send their local prototypes $\mathbf{p}_m$ along with their domain identifiers to the server. For each class $c$ and domain $d$, the server collects all prototypes from clients associated with domain $d$, forming a set $\{\mathbf{p}_j^{(c,d)}\}_{j=1}^{N_{c,d}}$, where $j$ indexes the clients that provided a prototype for same $(c,d)$ pair as shown in Step (2.1) of Fig.~\ref{system}. Here $N_{c,d}$ denotes the number of client prototypes available for class $c$ and domain $d$. We construct the domain-specific global prototype using a tailored cosine similarity-weighted fusion mechanism rather than the simple averaging direction. This mechanism assigns higher weights to prototypes that are more semantically consistent with others in the same domain, thus achieving more robust global domain-specific prototypes. Formally, the attention score $\alpha_j$ for each local prototype is computed as:
\begin{equation}
    S_j = \sum_{k \neq j} \cos(\mathbf{p}_j^{(c,d)}, \mathbf{p}_k^{(c,d)}), \quad
    \alpha_j = \frac{\exp(S_j / \tau_{agg})}{\sum_{l=1}^{N_{c,d}} \exp(S_l / \tau_{agg})},
\label{eq:attention}
\end{equation}
where $S_j$ denotes the consistency score computed against the other prototypes in the same set $\{\mathbf{p}_k^{(c,d)}\}_{k=1}^{N_{c,d}}$ (i.e., same class $c$ and same domain $d$). Here, $\tau_{agg}$ is a temperature parameter controlling the sharpness of the weighting distribution. The domain-specific prototype is then aggregated as:
% \begin{equation}
%     \mathbf{P}^{(c,d)} = \sum_{j=1}^{N_{c,d}} \alpha_j \cdot \mathbf{p}_j^{(c,d)} \in \mathbb{R}^{C \times I \times D}.
% \end{equation}
% \begin{equation}
% \mathbf{P} =
% \begin{bmatrix}
% \mathbf{P}^{(1,1)} & \mathbf{P}^{(2,1)} & \cdots & \mathbf{P}^{(C,1)} \\
% \mathbf{P}^{(1,2)} & \mathbf{P}^{(2,2)} & \cdots & \mathbf{P}^{(C,2)} \\
% \vdots             & \vdots             & \ddots & \vdots \\
% \mathbf{P}^{(1,D)} & \mathbf{P}^{(2,D)} & \cdots & \mathbf{P}^{(C,D)}
% \end{bmatrix}
% \in \mathbb{R}^{D \times C \times I},
% \end{equation}

\begin{equation}
\begin{aligned}
\mathbf{P}^{(c,d)} &= \sum_{j=1}^{N_{c,d}} \alpha_j \cdot \mathbf{p}_j^{(c,d)}\\
\mathbf{P} &= 
\begin{bmatrix}
\mathbf{P}^{(1,1)} & \mathbf{P}^{(2,1)} & \cdots & \mathbf{P}^{(C,1)} \\
\mathbf{P}^{(1,2)} & \mathbf{P}^{(2,2)} & \cdots & \mathbf{P}^{(C,2)} \\
\vdots             & \vdots             & \ddots & \vdots \\
\mathbf{P}^{(1,D)} & \mathbf{P}^{(2,D)} & \cdots & \mathbf{P}^{(C,D)} 
\end{bmatrix}  \in \mathbb{R}^{D \times C \times I} 
,
\end{aligned}
\label{eq:domain_proto_agg}
\end{equation}
where $\mathbf{P}^{(c,d)}$ denotes the domain-specific global prototype corresponding to class $c$ in domain $d$. The complete prototype tensor $\mathbf{P} \in \mathbb{R}^{D \times C \times I}$ arranges these vectors such that each row corresponds to a domain, each column to a class, and each entry stores a semantic feature embedding of dimension $I$.

\subsection{Dual Prototype Alignment Strategy}
Existing domain‐agnostic alignment forces every client to match the same global prototypes, overlooking the distinct feature distributions that arise under domain shift. To overcome this, we leverage our Domain‐Specific Global Prototypes by segmenting them into two groups according to the client domain: intra-domain and cross-domain prototypes. Subsequently, we design the tailored components that preserve each client's domain identity while modeling cross-domain relationships.
\subsubsection{Domain-Consistent Prototype Alignment}
Aligning local features to general global prototypes without considering domain-specific characteristics can lead to semantic inconsistency. To mitigate this, we first design a domain-consistent alignment that leverages the \textit{intra-domain prototypes}-those corresponding to the client's own domain. This ensures that each client aligns with semantically coherent while preserving domain-specific characteristics. For a given sample $(x_i^m, y_i^m)$ from client $m$ with domain label $d_m$, we extract the feature vector $z_i = f_\theta(x_i^m) \in \mathbb{R}^I$, and align it with the corresponding domain-specific prototype $\mathbf{P}^{(c, d_m)}$. Specifically, we propose the Domain-Consistent Prototype Alignment (DPA) between the local features and intra-domain prototypes as follows:
\begin{equation}
\mathcal{L}_{\text{DPA}} = \sum_{c=1}^{C} \left(1 - \cos\left(z^c_i, \mathbf{P}^{(c, d_m)}\right)\right),
\label{eq:dpa}
\end{equation}
where $z^c_i$ denotes the local features of class $c$. Cosine similarity is employed to emphasize the angular alignment between features and prototypes, rendering the loss more robust to scale variations across domains. This formulation encourages each local feature to align with its class-specific, domain-relevant prototype, thereby promoting semantic consistency and stabilizing training under domain shift.

\subsubsection{Cross-Domain Prototype Contrastive Learning}

Focusing only on intra-domain alignment may limit the model’s ability to generalize across domains. To overcome this limitation, we use contrastive learning to pull features toward same-class prototypes from other domains and push them away from different classes. By utilizing the prototypes from cross-domain group, we expose the model to diverse representations of each class and encourage the learning of domain-invariant semantics. Given a local feature \(z_i = f_\theta(x_i^m)\) on client \(m\) (domain \(d_m\)), we define its cross-domain positive and negative prototype sets as:
\begin{equation}
\begin{aligned}
\mathcal{P}_i^+ &= \bigl\{\,\mathbf{P}^{(y_i^m, d')}\;\big|\;d'\neq d_m\bigr\}, \\[6pt]
\mathcal{P}_i^- &= \bigl\{\,\mathbf{P}^{(c', d')}\;\big|\;c'\neq y_i^m,\;d'\neq d_m\bigr\}.
\end{aligned}
\end{equation}
Then, we enforce the instance local feature to be similar to the positive cross-domain  prototypes and dissimilar to the negative cross-domain prototypes. The Cross-Domain Prototype Contrastive Learning (CPCL) is presented as follows:
\begin{equation}
\mathcal{L}_{\mathrm{CPCL}}
= -\frac{1}{B}\sum_{i=1}^B
\log
\frac{
  \displaystyle\sum_{p\in\mathcal{P}_i^+}\exp\!\bigl(\cos(z_i,p)/\tau_{cross}\bigr)
}{
  \displaystyle\sum_{p\in\mathcal{P}_i^+\cup\mathcal{P}_i^-}\exp\!\bigl(\cos(z_i,p)/\tau_{cross}\bigr)
},
\label{eq:cpcl}
\end{equation}
where \(\tau_{cross}\) is a temperature hyperparameter, $B$ denotes the local batch size. The CPCL encourages model to capture semantic similarity in different domains, thereby improving its ability to generalize across heterogeneous client domains. Simultaneously, by maximizing the distance against negative class prototypes, the CPCL sharpens class-level discrimination, ensuring that representations remain well-separated even under domain shifts. Besides, we construct the CrossEntropy~\cite{de2005tutorial} loss to train the local model with logits output $s_i=g_\phi(z_i)$ to peform classification task as follows:
\begin{equation}
\mathcal{L}_{\mathrm{CE}} = -\mathbf{1}_{y_i} \log \left( \sigma(s_i) \right),
\label{eq:ce}
\end{equation}
where $\sigma$ denotes softmax. Finally, our overall objective is formulated as follows:
\begin{equation}
\mathcal{L} = \mathcal{L}_{\mathrm{CE}} + \lambda_1\mathcal{L}_{\mathrm{DPA}} + \lambda_2\mathcal{L}_{\mathrm{CPCL}},
\label{eq:total}
\end{equation}
\begin{algorithm}[t]
    \caption{\textbf{FedDAP}}
    \label{alg:algorithm}
    \textbf{Input}: Communication rounds R, local epochs E, number of clients M, local dataset $\mathcal{D}_m = \{(x_i^m, y_i^m)\}_{i=1}^{N_m}$.\\
    % \textbf{Parameter}: Optional list of parameters\\  
    \textbf{Output}: Global model $\Theta_t$
    \begin{algorithmic}[1] %[1] enables line numbers
    \STATE \textbf{Server Execution:} 
    \FOR{$t=0,\dots,{T-1}$}
    \FOR{$m=0,\dots,{M-1}$} 
    \item  $\Theta^m_t,\mathbf{p}_{m}\leftarrow\textbf{LocalUpdate}(\Theta_t,\mathbf{P}^{c,d})$
    \ENDFOR
    
    \STATE  Conduct the domain-specific global prototype aggregation in Sec.~\hyperref[proto_agg]{\textcolor{red}{\ref*{proto_agg}}}.

    \STATE \textcolor{steelblue}{/* Global model update */}
    
    $\Theta_{t+1}\leftarrow\sum_{m=1}^M\frac{|\mathcal{D}_m|}{|\mathcal{D}|}\Theta_t^m$
    \ENDFOR
    \STATE \textbf{Client Execution:}
    \STATE \textbf{LocalUpdate}($\Theta_t,\mathbf{P}^{c,d}$):
    \FOR{$e=0,\dots,E$}
    \FOR{each batch $\in \mathcal{D}_m = \{(x_i^m, y_i^m)\}_{i=1}^{N_m}$}
    \STATE Update $\Theta_t$ using $\mathbf{P}^{c,d}$ by Eq.~\hyperref[eq:total]{\textcolor{red}{~\ref*{eq:total}}}.
    \ENDFOR
    \ENDFOR
    \STATE  $\mathbf{p}_m^{(c, d)} = \frac{1}{|\mathcal{D}_m^{(c)}|} \sum_{(x_i^m, y_i^m) \in \mathcal{D}_m^{(c)}} f_\theta(x_i^m)$
    \STATE \textbf{return} $\Theta^m_t,\mathbf{p}_{m}^{c,d}$      
    \end{algorithmic}
\label{mainalg}
\end{algorithm}
where $\lambda_1$ and $\lambda_2$ balance the influence of the two prototype-based losses. Our design enables the model to benefit from both intra-domain stability and cross-domain generalization, yielding domain-aware and semantically meaningful representations in federated learning under domain shift. We illustrate the main flow in Alg.~\ref{mainalg}.
\section{Experiment}
\label{sec:formatting}

\subsection{Experimental Setup}

\begin{table*}[ht]
\centering
\resizebox{\textwidth}{!}{%
\begin{tabular}{l c c c c c c | c c | c c c c | c c}
\toprule
\multirow{2}{*}{\textbf{Method}}
  & \multicolumn{8}{c|}{\textbf{DomainNet}} & \multicolumn{6}{c}{\textbf{Office-10}} \\
\cmidrule(lr){2-9}\cmidrule(lr){10-15}
  & Clipart & Infograph & Painting & Quickdraw & Real & Sketch & Avg & $\Delta$
  & Caltech & Amazon & Webcam & DSLR & Avg & $\Delta$ \\
\midrule
FedAvg    & 67.73 & 31.28 & 58.22 & 71.20 & 70.72 & 58.41 & 59.59 & --    
          & 67.95 & 78.63 & 49.31 & 34.00 & 57.47 & --    \\
FedProx   & 68.46 & 34.08 & 58.87 & 74.02 & 72.40 & 55.62 & 60.57 & +0.98
          & 66.32 & 75.77 & 55.42 & 42.33 & 59.96 & +2.49\\
MOON      & 68.11 & 33.81 & 58.19 & 72.26 & 72.63 & 55.87 & 60.15 & +0.56
          & 64.25 & 77.19 & 55.15 & 43.00 & 59.89 & +2.42\\
COPA      & 68.84 & 34.24 & 60.45 & 76.94 & 73.04 & 54.65 & 61.36 & +1.77
          & 70.82 & 80.21 & 51.72 & 55.33 & 64.50 & +7.03\\
FedGA     & 67.20 & 33.29 & 60.81 & 74.20 & 72.34 & 58.19 & 61.00& +1.41
          & 65.27 & 76.11 & 57.24 & 51.33 & 62.49 & +5.02\\
FedProto  & 68.00 & 32.59 & 59.58 & 72.48 & 71.70 & 58.73 & 60.51 & +0.92
          & 66.30 & 78.52 & 58.62 & 55.33 & 64.69 & +7.22\\
FPL       & 68.68 & 35.76 & 60.39 & 74.52 & 72.22 & 61.48 & 62.17 & +2.58
          & 72.25 & 78.23 & 66.91 & 52.67 & 67.52 & +10.05\\
FedPLVM   & 69.40 & 34.15 & 60.81 & 75.37 & 73.17 & 59.85 & 62.22 & +2.63
          & 69.39 & 80.41 & 63.95 & 61.33 & 68.77 & +11.30\\

FedRDN
          & 65.83    & 35.31    & 56.60    & 77.94    & 74.67 & 55.69 & 61.01    & +1.42
          & 62.95    & 73.71    & 54.83    & 70.67    & 65.54   & +8.07      \\
\midrule          
\textbf{FedDAP}
          & 72.11 & 35.03 & 64.50 & 80.38 & 75.64 & 63.52 & \textbf{65.20} & \textbf{+5.61}
          & 73.75 & 82.42 & 68.62 & 65.33 & \textbf{72.53} & \textbf{+15.06} \\
\bottomrule
\end{tabular}%
}
\caption{Comparison on DomainNet and Office-10 datasets. “Avg” denotes the mean accuracy across all domains, and $\Delta$ indicates improvement over the FedAvg baseline.}
\label{comparison_domainnet_office}
\end{table*}

% \begin{table}[t]
% \centering
% \resizebox{\columnwidth}{!}{%
% \begin{tabular}{l c c c c|c c}
% \toprule
% \multirow{2}{*}{\textbf{Method}} 
%   & \multicolumn{6}{c}{\textbf{PACS}} \\ 
% \cmidrule(lr){2-7}
%   & Photo & Art & Cartoon & Sketch & Avg & $\Delta$ \\
% \midrule
% FedAvg    & 80.89 & 68.56 & 72.25 & 86.59 & 77.07 & –    \\
% FedProx   & 81.23 & 66.35 & 74.83 & 87.97 & 77.60 & +0.53\\
% MOON      & 83.47 & 72.69 & 74.55 & 91.66 & 80.59 & +3.52\\
% COPA      & 82.46 & 68.55 & 72.95 & 88.16 & 78.03 & +0.96\\
% FedGA     & 84.51 & 67.80 & 73.53 & 90.65 & 79.13 & +2.06\\
% FedProto  & 88.49 & 71.70 & 74.54 & 89.09 & 80.95 & +3.88\\
% FPL       & 86.72 & 70.96 & 75.52 & 91.22 & 81.11 & +4.04\\
% FedPLVM   & 88.44 & 71.22 & 75.84 & 92.73 & 82.06 & +4.99\\
% \midrule
% \textbf{FedDAP}
%           & 90.43 & 75.28 & 77.64 & 95.16 & \textbf{84.63} & \textbf{+7.56} \\
% \bottomrule
% \end{tabular}%
% }
% \caption{Comparison on the PACS dataset. “Avg” denotes the mean accuracy across all domains, and $\Delta$ indicates improvement over the FedAvg baseline.}
% \label{comparison_pacs}
% \end{table}

\begin{table}[t]
\centering
\resizebox{8cm}{!}{%
\begin{tabular}{l c c c c|c c}
\toprule
\multirow{2}{*}{\textbf{Method}} 
  & \multicolumn{6}{c}{\textbf{PACS}} \\ 
\cmidrule(lr){2-7}
  & Photo & Art & Cartoon & Sketch & Avg & $\Delta$ \\
\midrule
FedAvg    & 80.89 & 68.56 & 72.25 & 86.59 & 77.07 & --    \\
FedProx   & 81.23 & 66.35 & 74.83 & 87.97 & 77.60 & +0.53\\
MOON      & 83.47 & 72.69 & 74.55 & 91.66 & 80.59 & +3.52\\
COPA      & 82.46 & 68.55 & 72.95 & 88.16 & 78.03 & +0.96\\
FedGA     & 84.51 & 67.80 & 73.53 & 90.65 & 79.13 & +2.06\\
FedProto  & 88.49 & 71.70 & 74.54 & 89.09 & 80.95 & +3.88\\
FPL       & 86.72 & 70.96 & 75.52 & 91.22 & 81.11 & +4.04\\
FedPLVM   & 88.44 & 71.22 & 75.84 & 92.73 & 82.06 & +4.99\\
FedRDN
          & 85.29    & 76.79    & 77.07    & 93.51    & 83.17    & +6.10    \\
\midrule
\textbf{FedDAP}
          & 90.43 & 75.28 & 77.64 & 95.16 & \textbf{84.63} & \textbf{+7.56} \\
\bottomrule
\end{tabular}%
}
\caption{Comparison on the PACS dataset. “Avg” denotes the mean accuracy across all domains, and $\Delta$ indicates improvement over the FedAvg baseline.}
\label{comparison_pacs}
\end{table}
\noindent \textbf{Datasets.} We evaluate our method on three multi-domain datasets: DomainNet~\cite{peng2019moment}, Office-10~\cite{gong2012geodesic}, and PACS~\cite{li2017deeper}. \textbf{DomainNet} dataset contains six domains, including Clipart, Infograph, Painting, Quickdraw, Real, and Sketch. We follow~\cite{zhou2023fedfa} to select $10$ classes from $345$ classes of original dataset. \textbf{Office-10} consists of four domains: Caltech, Amazon, Webcam, and DSLR, covering $10$ shared object categories. The \textbf{PACS} dataset includes images from four distinct styles: Photo, Art Painting, Cartoon, and Sketch, spanning $7$ object classes. To simulate the domain shift scenario, we conduct the federated learning system with $10$ clients for Office-10 and PACS dataset, $20$ clients for DomainNet dataset, with the client allocation spanning across different domains. The client-to-domain allocations are as follows: for DomainNet: Clipart ($3$), Infograph ($1$), Painting ($4$), Quickdraw ($6$), Real ($4$), Sketch ($2$); for Office-10: Caltech ($3$), Amazon ($2$), Webcam ($1$), DSLR ($4$); and for PACS: Photo ($3$), Art Painting ($2$), Cartoon ($1$), Sketch ($4$). For each client, we sampled data from the these domains according to the difficulty and size of each dataset, using sampling rates of $10\%$ for DomainNet, $20\%$ for Office-10, and $30\%$ for PACS.

\noindent \textbf{Baselines.} We compare our proposed methods against several SOTA FL approaches: typical FL methods: \textbf{FedAvg}~(AISTATS'17)~\cite{mcmahan2017communication}, \textbf{FedProx}~(MLsys'21)~\cite{li2020federated_1}, \textbf{MOON}~(CVPR'21)~\cite{li2021model}, Federated Domain Generalization methods: \textbf{COPA}~(ICCV'21)~\cite{wu2021collaborative} and \textbf{FedGA}~(CVPR'23)~\cite{zhang2023federated}, as well as prototype-based FL methods: \textbf{FedProto}~(AAAI'22)~\cite{tan2022fedproto} (with parameter averaging), ~\textbf{FPL}~(CVPR'23)~\cite{huang2023rethinking},~\textbf{FedPLVM}~(NeurIPS'24)~\cite{wang2024taming}, and the latest domain shift FL method: \textbf{FedRDN}~(CVPR'25)~\cite{yan2025simple}. 

\begin{figure*}[]
    \centering
    \begin{subfigure}{0.3\textwidth}
        \centering
        \includegraphics[width=\linewidth]{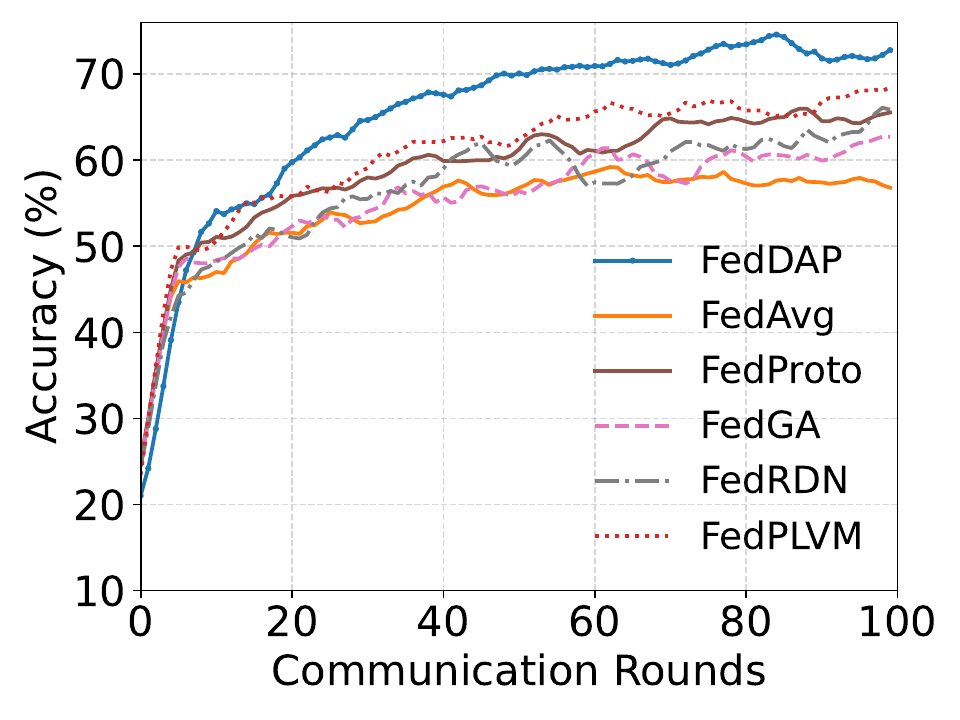}
        \caption{Office-10}
    \end{subfigure}%
    \hfill
      \begin{subfigure}{0.3\textwidth}
        \centering
        \includegraphics[width=\linewidth]{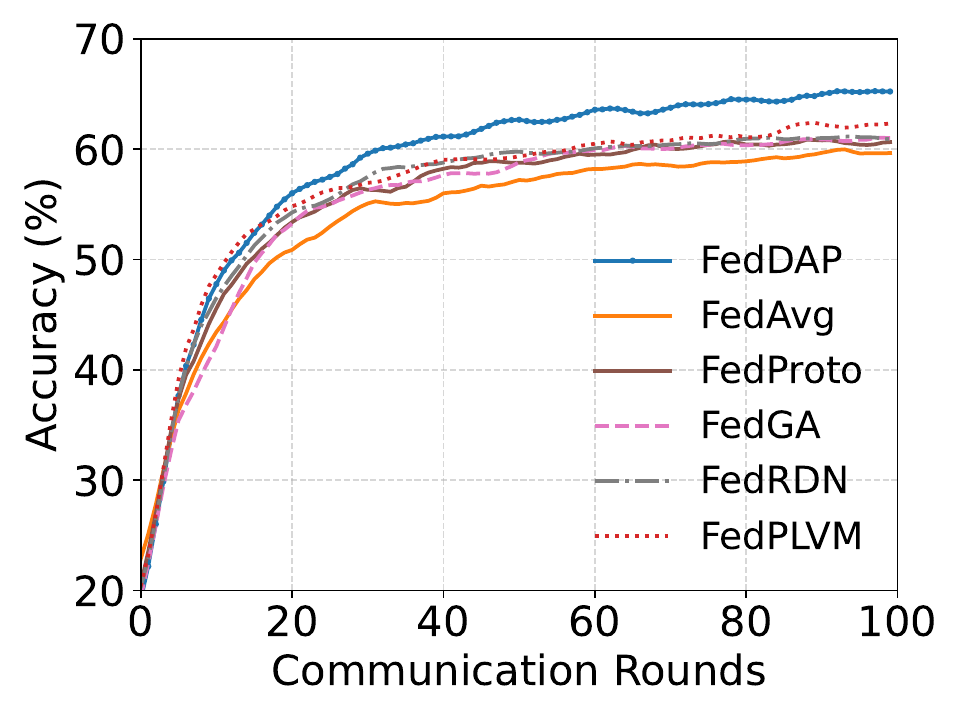}
        \caption{DomainNet}
    \end{subfigure}%
    \hfill
    \begin{subfigure}{0.30\textwidth}
        \centering
        \includegraphics[width=\linewidth]{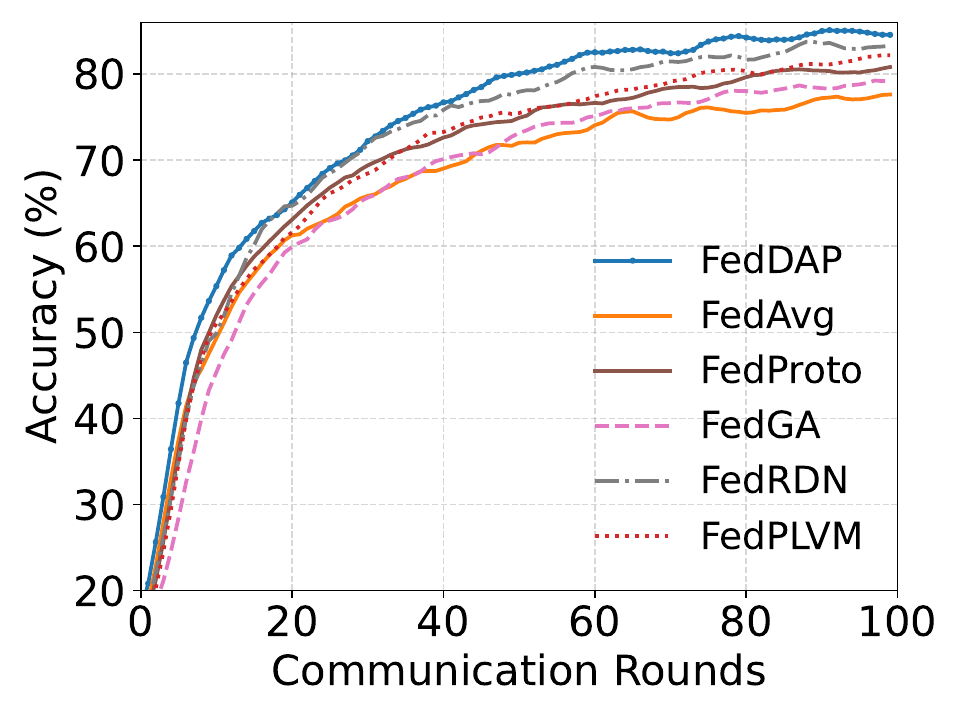}
        \caption{PACS}
    \end{subfigure}
    \caption{Visualization of training curves of average test accuracy on three datasets under the domain shift setting.}
    \label{curve}
\end{figure*}

\begin{figure}[]
  \centering
  % (a) DBLP: FedProto vs Ours
  \begin{subfigure}[b]{0.85\columnwidth}
    \centering
    \includegraphics[width=0.50\linewidth]{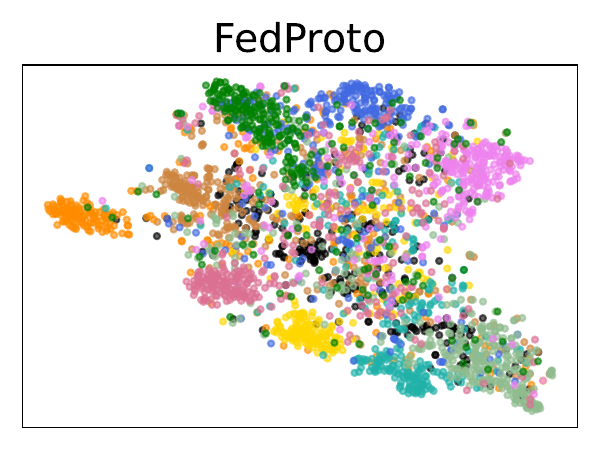}%
    \hfill
    \includegraphics[width=0.50\linewidth]{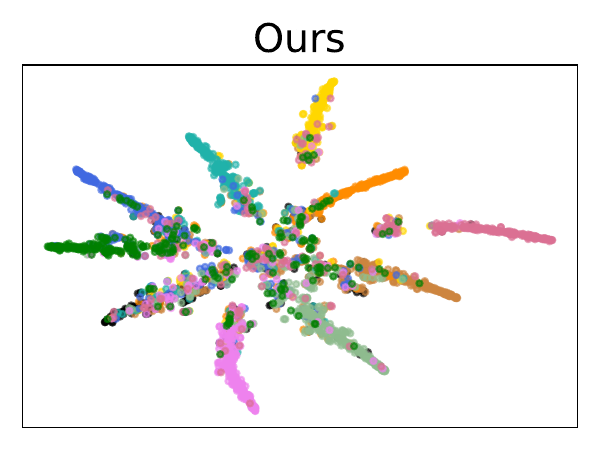}%
    \caption{DomainNet}
    \label{fig:tsne_digit}
  \end{subfigure}%
  \hfill%
  % (b) ACM: FedProto vs Ours
  \begin{subfigure}[b]{0.85\columnwidth}
    \centering
    \includegraphics[width=0.50\linewidth]{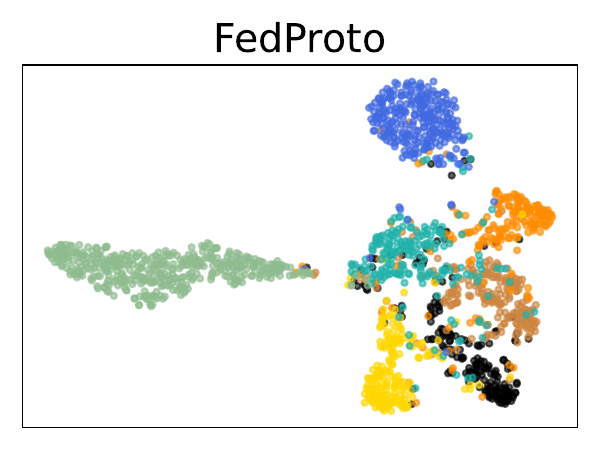}%
    \hfill
    \includegraphics[width=0.50\linewidth]{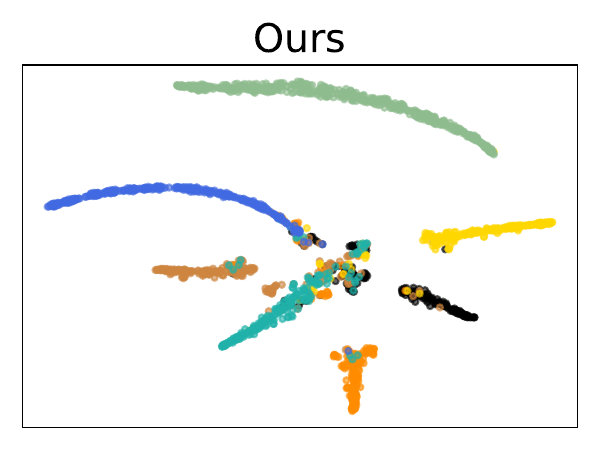}%
    \caption{PACS}
    \label{fig:tsne_pacs}
  \end{subfigure}

  \caption{%
 t-SNE visualization of learned features on the DomainNet and PACS benchmarks. In each subfigure, the left panel shows FedProto and the right panel shows our method.
  }
  \label{fig:tsne}
\end{figure}

\noindent \textbf{Implementation Details.} We conduct experiments using ResNet-10~\cite{he2016deep} as the backbone for the DomainNet and Office-10 datasets, and ResNet-18 for the PACS dataset. The total number of communication rounds is set to 100, with each client performing $10$ local epochs per round. We employ a batch size of $32$ for DomainNet and Office-10, and $16$ for PACS dataset. Top-1 accuracy is used as the evaluation metric. Each experiment is repeated three times, and we report the mean accuracy over the last $5$ communication rounds.
\subsection{Performance Comparison with SOTA Methods}
\noindent \textbf{Performance Comparison.} We evaluate FedDAP against multiple baselines on three multi-domain datasets: DomainNet, Office-10, and PACS. As shown in Tables~\ref{comparison_domainnet_office} and~\ref{comparison_pacs}, FedDAP consistently achieves the highest average accuracy across all datasets. Compared to the standard FedAvg baseline, it outperforms by a large margin: $+5.61\%$ on DomainNet, $+15.06\%$ on Office-10, and $+6.86\%$ on PACS. It also surpasses recent prototype-based and federated domain generalization methods such as FedProto, FedPLVM, and FedGA, demonstrating strong generalization in both mild and severe domain shift scenarios. These results confirm the effectiveness of our domain-aware design. By leveraging domain-specific global prototypes and incorporating both intra-domain alignment and cross-domain contrastive learning, FedDAP successfully captures discriminative semantics while preserving domain consistency. Together, these components enable FedDAP to generalize well across heterogeneous domains and maintain strong class separability, leading to state-of-the-art performance across all benchmarks. To complement the quantitative gains above, we provide a qualitative assessment of feature separability via t-SNE visualization in Fig.~\ref{fig:tsne}. As the figure shows, FedProto's features tend to be more diffuse with residual overlap, while our FedDAP produce uniformly tighter and well-separated clusters. This behavior illustrates that FedDAP not only maintains the domain consistency but also enhances the semantic discrimination.

\noindent \textbf{Convergence.} As shown in Fig.~\ref{curve}, FedDAP achieves the faster convergence across all three datsets. On different datasets, our proposed framework clearly outperforms all baselines by a significant margin, achieving the higher accuracy in fewer communication rounds. These results demonstrate that our method not only improves the final average performance but also enhances the communication efficiency under domain shift.

\begin{table}[t]
\centering
\resizebox{\columnwidth}{!}{%
\begin{tabular}{c c | c c | c c | c c}
\toprule
\multirow{2}{*}{$\mathcal{L}_{\mathrm{DPA}}$} &
\multirow{2}{*}{$\mathcal{L}_{\mathrm{CPCL}}$} &
\multicolumn{2}{c|}{\textbf{DomainNet}} &
\multicolumn{2}{c|}{\textbf{Office-10}} &
\multicolumn{2}{c}{\textbf{PACS}} \\
\cmidrule(lr){3-8}
& & Avg & $\Delta$ & Avg & $\Delta$ & Avg & $\Delta$ \\
\midrule
\ding{55} & \ding{55} & 59.59 & --    & 57.47 & --     & 77.07 & --    \\
\ding{55} & \ding{51} & 62.86 & +3.27 & 62.18 & +4.71  & 81.87 & +4.80 \\
\ding{51} & \ding{55} & 62.61 & +3.02 & 68.53 & +11.06 & 78.74 & +1.67 \\
\ding{51} & \ding{51} & \textbf{65.20} & \textbf{+5.61} & \textbf{72.53} & \textbf{+15.06} & \textbf{84.63} & \textbf{+7.56} \\
\bottomrule
\end{tabular}%
}
\caption{Ablation study on key components of FedDAP across DomainNet, Office-10, and PACS datasets.}
\label{ablation_key_digit_office}
\end{table}

% \begin{table}[]
% \centering
% \resizebox{8.5cm}{!}{%
% \begin{tabular}{|c|cc|cc|cc|}
% \hline
% \multirow{2}{*}{\begin{tabular}[c]{@{}c@{}}Domain-Specific\\ Prototypes Aggregation\end{tabular}} & \multicolumn{2}{c|}{Digits} & \multicolumn{2}{c|}{Office-10} & \multicolumn{2}{c|}{PACS} \\
%                                                                                                   & Avg       & $\triangle$     & Avg         & $\triangle$      & Avg       & $\triangle$    \\ \hline
% Averaging                                                                                         & 81.34     & -               & 71.49       & -                & 83.22     & -             \\
% Ours                                                                                              & \textbf{83.11}     & \textbf{+1.77}           & \textbf{72.53}       & \textbf{+1.04}            & \textbf{84.63}     & \textbf{+1.41}         \\ \hline
% \end{tabular}
% }%
% \caption{Analysis of FedDAP with different domain-specific prototype aggregation methods.}
% \label{prototype_agg_analysis}
%\end{table}

\begin{table}[t]
\centering
\resizebox{\columnwidth}{!}{%
\begin{tabular}{l cc cc cc}
\toprule
\multirow{2}{*}{\textbf{Aggregation Method}} 
  & \multicolumn{2}{c}{\textbf{DomainNet}} 
  & \multicolumn{2}{c}{\textbf{Office-10}} 
  & \multicolumn{2}{c}{\textbf{PACS}} \\
\cmidrule(lr){2-3} \cmidrule(lr){4-5} \cmidrule(lr){6-7}
  & Avg & $\Delta$ 
  & Avg & $\Delta$ 
  & Avg & $\Delta$ \\
\midrule
Averaging & 64.15 & –      & 71.49 & –      & 83.22 & –     \\
Ours      & \textbf{65.20} & \textbf{+1.05} 
          & \textbf{72.53} & \textbf{+1.04} 
          & \textbf{84.63} & \textbf{+1.41} \\
\bottomrule
\end{tabular}%
}
\caption{Comparison of domain-specific prototype aggregation methods. “Avg” denotes mean accuracy and “$\Delta$” the improvement over uniform averaging.}
\label{prototype_agg_analysis}
\end{table}

\begin{table*}[]
\centering
\resizebox{\textwidth}{!}{%
\begin{tabular}{l c c c c c c | c c | c c c c | c c}
\toprule
\multirow{2}{*}{\textbf{Method}}
  & \multicolumn{8}{c|}{\textbf{DomainNet}} & \multicolumn{6}{c}{\textbf{Office-10}} \\
\cmidrule(lr){2-9}\cmidrule(lr){10-15}
  & $\rightarrow$ Clipart & $\rightarrow$ Infograph & $\rightarrow$ Painting & $\rightarrow$ Quickdraw & $\rightarrow$ Real & $\rightarrow$ Sketch & Avg & $\Delta$
  & $\rightarrow$ Caltech & $\rightarrow$ Amazon & $\rightarrow$ Webcam & $\rightarrow$ DSLR & Avg & $\Delta$ \\
\midrule
FedAvg    & 40.25 &46.91 & 45.83 & 43.68   &  42.36  &44.39 &  43.90& --    
          &  43.54   & 65.81   &  61.64   & 70.04 &60.26  & --    \\
COPA      &  43.09  & 47.52  &  47.21 & 41.86 &  44.12 &   43.48 & 44.55 & +1.01
          & 49.75 &  69.36  &   60.35  & 71.05 & 62.63 & +2.37\\
FedGA     & 44.48 & 46.40 & 45.45  &  43.79 &44.60  &43.06  &44.63 &+0.73 
          & 45.88 & 65.19  &  62.61 & 69.55 & 60.81 & +0.55\\
FedPLVM   &46.75 & 49.24 & 50.28 &  45.95 &46.76 & 46.86  & 47.64 & +3.74
          & 50.92 & 71.95 & 62.91 &  72.33 & 64.53 & +4.27\\

FedRDN
          &  40.15    & 47.21    & 47.89     &  41.39     & 43.48 & 45.32 & 44.24    & +0.34
          &  39.61    & 74.63       &64.71   &   68.94    & 61.97   & +1.71      \\
\midrule          
\textbf{FedDAP}
          & 49.25 & 52.18 & 51.52  &  46.63 &  50.04   & 47.52 & \textbf{49.52} & \textbf{+5.62}
          &  52.88 & 73.07  &  63.65 &  73.18 & \textbf{65.70} & \textbf{+5.44} \\
\bottomrule
\end{tabular}%
}
\caption{Comparison on DomainNet and Office-10 datasets under domain generalization setting. “Avg” denotes the mean accuracy across all domains, and $\Delta$ indicates improvement over the FedAvg baseline.}
\label{comparison_domainnet_office_domaingen}
\end{table*}

\subsection{Ablation Study on Key Components}
In Table~\ref{ablation_key_digit_office}, we isolate the two core components of FedDAP: \textbf{intra-domain prototype alignment} $\mathcal{L}_{DPA}$ and \textbf{cross-domain prototype contrastive learning} $\mathcal{L}_{CPCL}$ to analyze their individual contributions. Without both components, the method reduces to a FedAvg-style aggregator and struggle to maintain either domain consistency. Notably, $\mathcal{L}_{DPA}$ consistently improves the results by reinforcing prototype consistency within each domain, illustrating the importance of intra-domain alignment. On the other hand, $\mathcal{L}_{CPCL}$ achieve the even larger gains by explicitly discouraing confusion between different classes across domains, demonstrating the effectiveness of the cross-domain discrimination. Combining both components produces the best accuracy, illustrating that intra- and cross-domain prototypes are essential for robust generalization under domain shift.
\subsection{Analysis on Domain-Specific Prototypes}
Table~\ref{prototype_agg_analysis} compares our similarity-weighted fusion mechanism against simple averaging method for domain specific prototypes aggregation. While averaging method yields the solid performance, our aggregation consistently improves performance by $1.05\%$, $1.04\%$, and $1.44\%$ on DomainNet, Office-10, and PACS dataset, respectively. This demonstrates weighting prototypes according to the domain similarity more effectively captures the domain-consistent feature than uniform averaging, further enhancing FedDAP's generalization under domain shift. Notably, even with simple averaging in domain-aware global aggregation, the performance improves significantly-highlighting the effectiveness and informativeness of domain-specific prototypes.
\begin{figure}[]
    \centering
    % Row 1: temperatures
    \begin{subfigure}{0.48\linewidth}
        \centering
        \includegraphics[width=\linewidth]{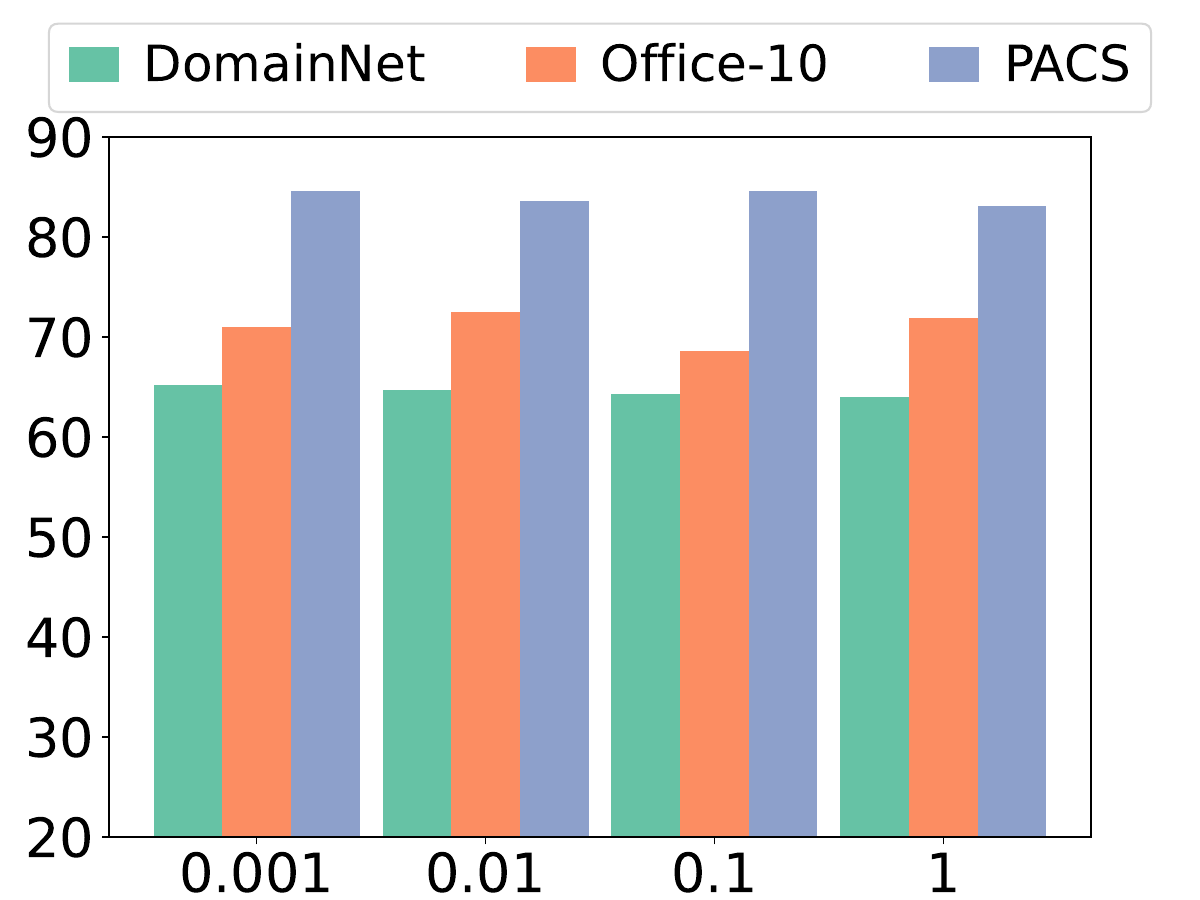}
        \caption{Temperature $\tau_{\text{agg}}$}
        \label{fig:temperature_agg}
    \end{subfigure}\hfill
    \begin{subfigure}{0.48\linewidth}
        \centering
        \includegraphics[width=\linewidth]{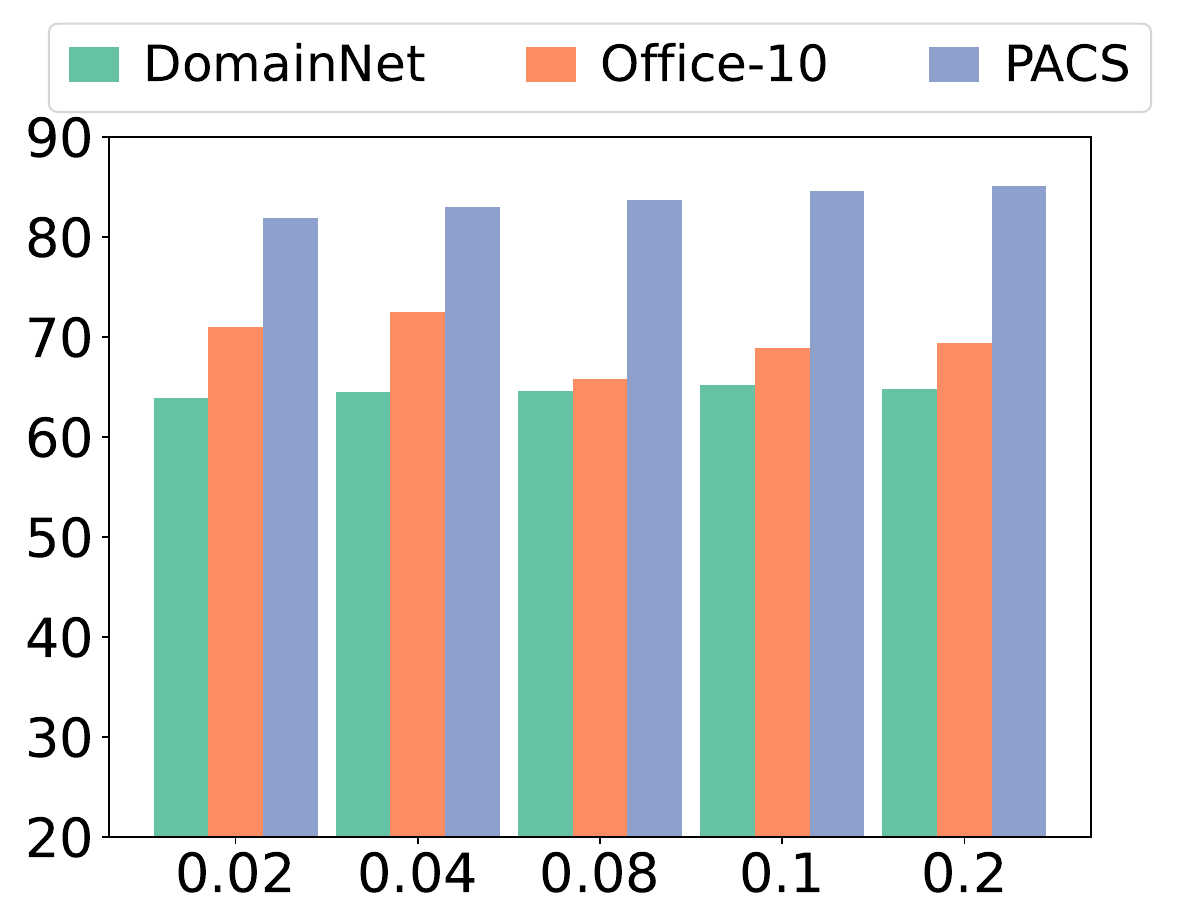}
        \caption{Temperature $\tau_{\text{cross}}$}
        \label{fig:temp_cross}
    \end{subfigure}

    \vspace{0.6em}

    % Row 2: lambdas
    \begin{subfigure}{0.48\linewidth}
        \centering
        \includegraphics[width=\linewidth]{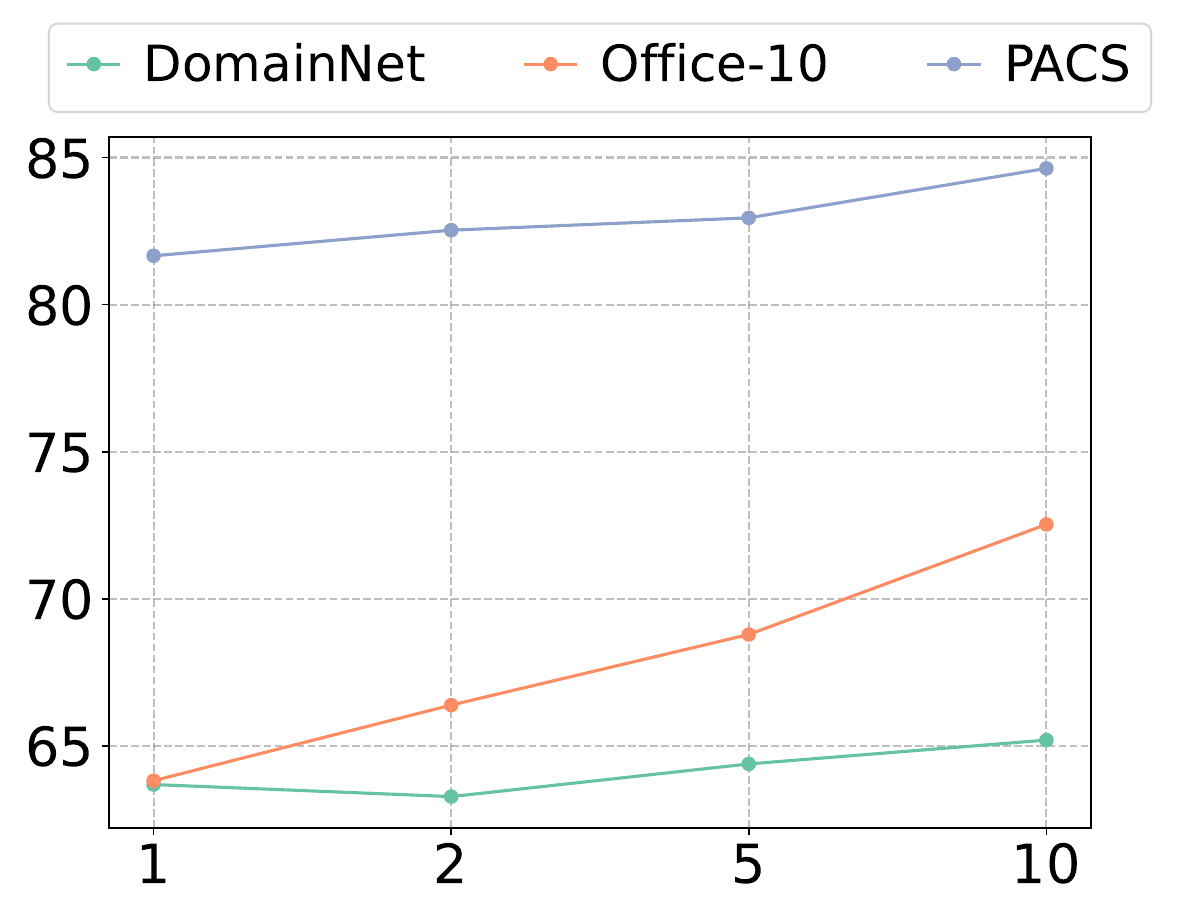}
        \caption{$\lambda_{1}$}
        \label{fig:lambda1}
    \end{subfigure}\hfill
    \begin{subfigure}{0.48\linewidth}
        \centering
        \includegraphics[width=\linewidth]{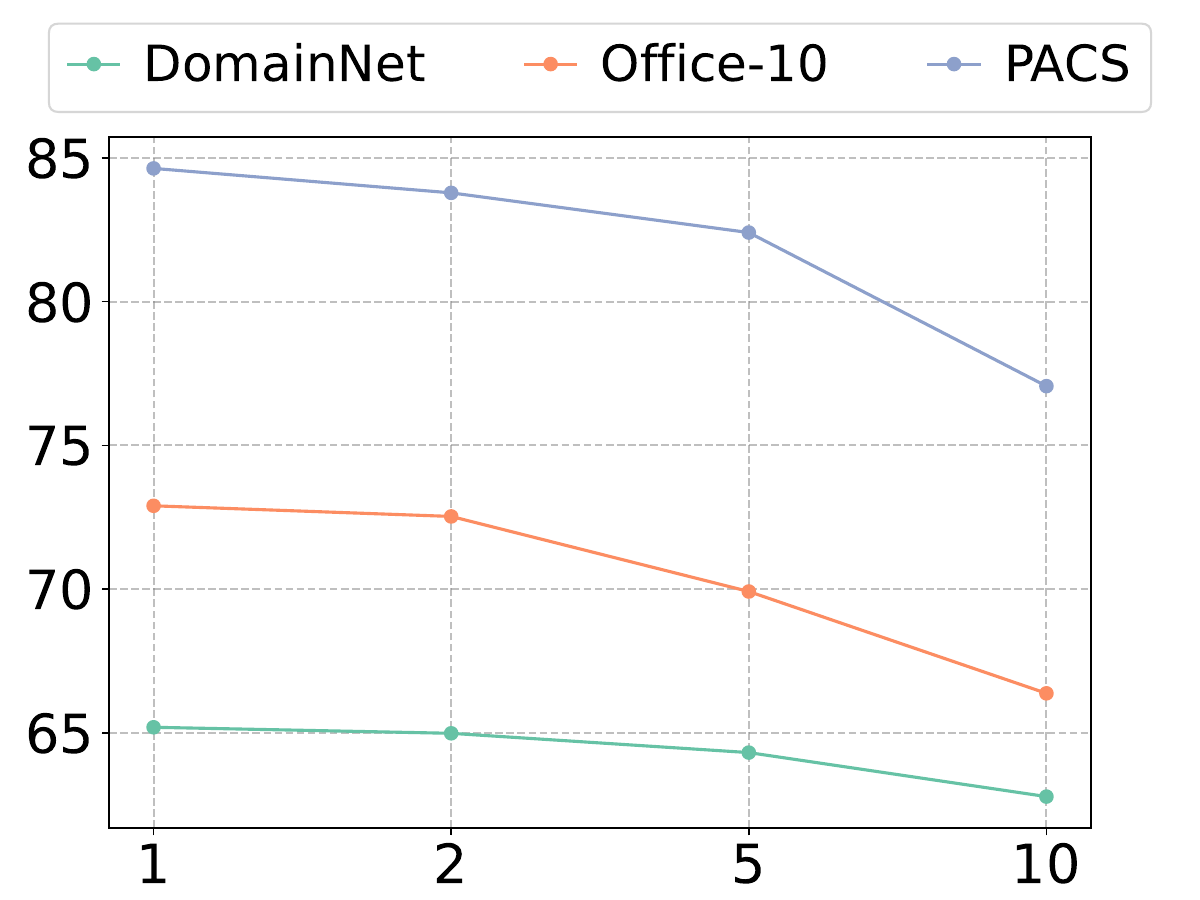}
        \caption{$\lambda_{2}$}
        \label{fig:lambda2}
    \end{subfigure}

    \caption{Analysis of FedDAP across all datasets with different hyper-parameters.}
    \label{fig:feddap_temps_hparams}
\end{figure}
\subsection{Cross-Domain Generalization Performance}
We evaluate our method under cross-domain generalization using a leave-one-domain-out protocol. For each dataset, one domain is held out for testing while the model is trained on remaining domains. As shown in Table~\ref{comparison_domainnet_office_domaingen}, FedDAP consistently outperforms other baselines on both DomainNet and Office-10 in terms of average target accuracy, indicating the superior robustness to domain shift and better generalization to unseen domains. By maintaining domain-specific global prototypes for each class, the model separates label semantics from domain style. This avoids the conflict that arises when all domains are forced to match a single prototype. In addition, our cross-domain alignment pulls same-class features across domains together while separating different classes, yielding tighter domain-aware clusters and reduced inter-domain scatter, thus improving robustness to domain shift and generalization to unseen domains.

% \begin{figure}[]
%   \centering

%   % (a) Digits (MNIST domain)
%   \begin{subfigure}[b]{\columnwidth}
%     \centering
%     \includegraphics[width=0.5\linewidth]{figure/protos_fedprotofl_digitsdomain0_2nd_paper.pdf}%
%     \hfill
%     \includegraphics[width=0.5\linewidth]{figure/protos_feddomainfl_digitsdomain0_2nd_paper.pdf}%
%     \caption{Digits (MNIST domain)}
%     \label{fig:tsne_digit}
%   \end{subfigure}

%   % \vspace{1em} % Add vertical spacing between rows if needed

%   % (b) PACS (Photo domain)
%   \begin{subfigure}[b]{\columnwidth}
%     \centering
%     \includegraphics[width=0.5\linewidth]{figure/protos_fedprotofl_pacsdomain0_2nd_paper.pdf}%
%     \hfill
%     \includegraphics[width=0.5\linewidth]{figure/protos_feddomainfl_pacsdomain0_2nd_paper.pdf}%
%     \caption{PACS (Photo domain)}
%     \label{fig:tsne_pacs}
%   \end{subfigure}

%   \caption{t-SNE visualization of learned features on the Digits and PACS benchmarks. In each subfigure, the left panel shows FedProto and the right panel shows our method.}
%   \label{fig:tsne}
% \end{figure}

\subsection{Hyperparameter Analysis}
Fig.~\ref{fig:temperature_agg} and~\ref{fig:temp_cross} examine how the two temperature hyper-parameters, $\tau_{agg}$ and $\tau_{cross}$ affect FedDAP's performance on different datasets. For the aggregation temperature $\tau_{\text{agg}}$, Office-10 improves as $\tau_{\text{agg}}$ increases and stabilizes near $0.01$; DomainNet and PACS prefer smaller values, peaking at $\tau_{\text{agg}}{=}0.001$ and dropping slightly afterward.  For the contrastive temperature $\tau_{\text{cross}}$, Office-10 peaks around $0.04$ and then levels off, whereas DomainNet and PACS performance  improve with larger $\tau_{\text{cross}}$, suggesting that a higher contrastive temperature better tackles cross-domain generalization.
Figs~\ref{fig:lambda1} and \ref{fig:lambda2} highlight a trade-off between intra-domain consistency and cross-domain regularization. Increasing $\lambda_1$ consistently improves performance (best at $\lambda_1{=}10$) because stronger intra-domain alignment reduces within-domain inconsistency and stabilizes the domain-specific prototypes used as supervision. In contrast, $\lambda_2$ is most effective at a moderate value (best at $\lambda_2{=}1$): a small cross-domain contrastive term improves generalization, while overly large $\lambda_2$ can over-regularize cross-domain matching, weakening domain-specific structure and degrading accuracy.

\section{Conclusion}

In this paper, we introduce FedDAP, a domain‐aware prototype learning framework for federated learning under domain shift. Our approach incorporates two key components: the construction of domain-specific prototypes to preserve the domain semantics, and a dual alignment strategy that guides local training through intra-domain consistency enforcement and cross-domain contrastive learning. By integrating these components, FedDAP enforces both prototype alignment within the intra-domain and discrimination across domains, substantially enhancing the global model’s generalization. Experiments on multiple benchmarks demonstrate that FedDAP consistently outperforms state‐of‐the‐art prototype‐based and baseline FL methods with the large margins under domain shifts while maintaining robustness to hyperparameter variation. These results confirm the effectiveness of leveraging both domain‐specific and class‐aware prototypes to mitigate domain heterogeneity in federated learning.

\section*{Acknowledgement}
This work was supported by Institute of Information \& communications Technology Planning \& Evaluation(IITP) grant funded by the Korea government(MSIT) (No.2019-0-01287-005, Evolvable Deep Learning Model Generation Platform for Edge Computing), IITP-ITRC(Information Technology Research Center) grant funded by the Korea government(MSIT) (IITP-2025-RS-2023-00258649;50\%), NRF(National Research Foundation) grant funded by the Korea government(MSIT) (No. RS-2024-00352423), IITP grant funded by the MSIT (No. RS-2024-00509257, Global AI Frontier Lab) and Global-Learning \& Academic Research Institution for Master’s, PhD students, and Postdocs (G-LAMP) Program of the National Research Foundation of Korea (NRF), funded by the Ministry of Education (No. RS-2025-25442355). 
{
    \small
    \bibliographystyle{ieeenat_fullname}
    \bibliography{main}
}
\clearpage
\setcounter{page}{1}
\maketitlesupplementary

\section{More Details}
\subsection{Details of Baselines}
We consider the following baselines in this work, including standard federated learning methods as well as advanced approaches specifically designed to address domain shift problem.
\begin{itemize}
    \item \textbf{FedAvg}~\cite{mcmahan2017communication} is the vanilla FL method that update the global model by iteratively averaging the local trained models. 
    \item \textbf{FedProx}~\cite{li2020federated_1} introduces a proximal term in the local objective to reduce the client model drift and stabilize training under non-IID data. We set the $\mu=0.001$ on all datasets.
    \item \textbf{MOON}~\cite{li2021model} improves client-server consistency by using contrastive learning to align local model representations with the global model while pushing them away from the previous round local models, thereby mitigating client drift under non-IID data. We follow the original setting to tune $\mu$ from $\{0.5,1,5,10\}$.
    \item \textbf{COPA}~\cite{wu2021collaborative} introduces a decentralized learning framework that collaboratively optimizes local models and aggregates knowledge across nodes to improve domain generalization and adaptation without relying on a centralized server.
    \item \textbf{FedGA}~\cite{zhang2023federated} proposes a federated domain generalization framework that adjusts client updates based on estimated generalization gaps to improve robustness to unseen target domains. We follow the method to set the step size of GA method to $0.05$.
    \item \textbf{FedProto}~\cite{tan2022fedproto} introduces a framework that aligns clients through shared class prototypes, enabling effective collaboration across non-IID data. We set $\mu$ to $1$ for FedProto on all datasets.
    \item \textbf{FPL}~\cite{huang2023rethinking} proposes the clustering prototypes to achieve the unbiased prototypes under domain shift challenge. We set the contrastive temperature $\tau=0.02$ on all datasets.
    \item \textbf{FedPLVM}~\cite{wang2024taming} incorporates a dual-level prototype clustering approach and $\alpha$-sparsity prototype loss to tackle the domain shift challenge in FL. We follow the method to maintain default hyper-parameter values: $\tau=0.07$, $\alpha=0.25$, and $\lambda=100$ on all datasets.
    \item \textbf{FedRDN}~\cite{yan2025simple} introduces a plug-and-play data augmentation strategy that alleviates feature-distribution skew in federated learning and can be easily integrated into existing augmentation pipelines.
       
\end{itemize}
\section{Additional Results}
\subsection{Efficiency}
Table~\ref{tab:comm_flops_time} compares communication and computation costs on Office-10 dataset. 
Introducing domain-specific global prototypes increases only the \emph{download} traffic, since the server broadcasts one prototype per domain: $\textbf{Prototype}_{\mathrm{down}} = D \times \textbf{Prototype}_{\mathrm{up}}$, where $D$ is the number of domains. This additional payload is negligible compared with sending full model parameters. 
All methods have comparable computation, while our training time (32.84\,s) is lower than FedPLVM (37.28\,s) though higher than FedProto (21.20\,s). 
Despite the extra \emph{communication} and \emph{time} overhead, the accuracy gains of our method render the trade-off favorable.
 \begin{table}[]
\centering
\resizebox{\columnwidth}{!}{%
\begin{tabular}{l c c c c}
\toprule
\textbf{Method} &
\textbf{Prototype\(_{\mathrm{up}}\) (MB)} &
\textbf{Prototype\(_{\mathrm{down}}\) (MB)} &
\textbf{FLOPs\(\,\times 10^{10}\)} &
\textbf{Time\(_{\mathrm{train}}\) (s)} \\
\midrule
FedProto & 0.195 & 0.195 & 119.14 & 21.20 \\
FedPLVM  & 0.195 & 0.195 & 119.14 & 37.28 \\
FedDAP   & 0.195 & 0.781 & 119.14 & 32.84 \\
\bottomrule
\end{tabular}%
}
\caption{Comparison of communication and computation costs.}
\label{tab:comm_flops_time}
\end{table}

\begin{table}[]
\centering
\resizebox{\columnwidth}{!}{%
\begin{tabular}{l cc cc cc}
\toprule
\multirow{2}{*}{\textbf{Local Prototypes}} 
  & \multicolumn{2}{c}{\textbf{DomainNet}} 
  & \multicolumn{2}{c}{\textbf{Office-10}} 
  & \multicolumn{2}{c}{\textbf{PACS}} \\
\cmidrule(lr){2-3} \cmidrule(lr){4-5} \cmidrule(lr){6-7}
  & Avg & $\Delta$ 
  & Avg & $\Delta$ 
  & Avg & $\Delta$ \\
\midrule
\texttt{w} DP & 64.38 & –      & 71.51 & –      & 83.63 & –     \\
\texttt{w/o} DP      & \textbf{65.20} & \textbf{+0.82} 
          & \textbf{72.53} & \textbf{+1.02} 
          & \textbf{84.63} & \textbf{+1.00} \\
\bottomrule
\end{tabular}%
}
\caption{Impact of differential privacy on local prototypes across different datasets.}
\label{prototype_dp}
\end{table}
\begin{table*}[]
\centering
\resizebox{\textwidth}{!}{%
\begin{tabular}{l c c c c c c | c c | c c c c | c c}
\toprule
\multirow{2}{*}{\textbf{Method}} 
  & \multicolumn{8}{c|}{\textbf{DomainNet}} & \multicolumn{6}{c}{\textbf{Office-10}} \\
\cmidrule(lr){2-9}\cmidrule(lr){10-15}
  & Clipart & Infograph & Painting & Quickdraw & Real & Sketch & Avg & $\Delta$
  & Caltech & Amazon & Webcam & DSLR & Avg & $\Delta$ \\
\midrule
FedAvg    & 64.91 & 31.86 & 56.89 & 60.58 & 59.42 & 53.70 & 54.56 & --
          & 67.61 & 78.86 & 73.72 & 51.67 & 67.97 & -- \\
FedProx   & 63.47 & 34.05 & 55.76 & 58.68 & 65.49 & 52.97 & 55.07 & +0.51
          & 65.42 & 82.56 & 68.44 & 53.00 & 67.36 & -0.61 \\
MOON      & 65.64 & 30.52 & 54.37 & 57.62 & 67.02 & 51.85 & 54.28 & –0.28
          & 63.26 & 81.82 & 60.41 & 52.00 & 64.37 & -3.60 \\
COPA      & 63.72 & 32.36 & 57.93 & 63.85 & 61.86 & 52.62 & 55.39 & +0.83
          & 69.02 & 81.37 & 76.21 & 60.00 & 71.65 & +3.68 \\
FedGA     & 65.37 & 32.85 & 55.85 & 62.96 & 63.66 & 56.03 & 56.12 & +1.56
          & 65.56 & 81.05 & 74.14 & 56.33 & 69.27 & +1.30 \\
FedProto  & 64.98 & 31.31 & 55.18 & 64.80 & 66.30 & 53.73 & 56.05 & +1.49
          & 64.91 & 80.95 & 76.52 & 62.00 & 71.10 & +3.13 \\
FPL       & 66.13 & 33.44 & 58.03 & 63.78 & 69.75 & 56.85 & 57.98 & +2.49
          & 67.35 & 81.99 & 73.02 & 65.33 & 71.92 & +3.95 \\
FedPLVM   & 66.93 & 33.29 & 57.64 & 64.56 & 68.82 & 55.11 & 57.66 & +3.10
          & 69.06 & 81.66 & 72.58 & 67.00 & 72.58 & +4.61 \\
FedRDN
          & 65.62    & 32.47   & 55.88    & 64.50    & 65.13 & 54.50 & 56.35    & +1.79
          & 63.57   & 65.32    & 68.62   & 77.33   & 68.71   & +0.74      \\
\midrule
\textbf{FedDAP}
          & 70.59 & 33.96 & 62.62 & 69.76 & 70.18 & 62.28 & \textbf{61.57} & \textbf{+7.01}
          & 66.96 & 81.79 & 77.93 & 71.33 & \textbf{74.50} & \textbf{+6.53} \\
\bottomrule
\end{tabular}%
}
\caption{Comparison on DomainNet and Office-10 datasets under equal client allocation. “Avg” denotes the mean accuracy across all domains, and $\Delta$ indicates improvement over the FedAvg baseline.)}
\label{comparison_domainnet_office_equal}
\end{table*}
\subsection{Privacy preservation analysis}
Inspired by DBE~\cite{zhang2023eliminating}, we conducted a differential privacy (DP) analysis across different datasets. Specifically, we added Gaussian noise to local prototypes $\tilde{\mathbf{p}}_m^{(c, d)} = \mathbf{p}_m^{(c, d)} + q \cdot \mathcal{N}\bigl(0, s^2\bigr)$ with perturbation coefficient $q=0.1$ for the noise and a scale parameter $s=0.05$. As presented in Table~\ref{prototype_dp}, the results indicate that our method retains competitive performance even when subject to privacy-preserving noise.
\subsection{Performance comparison on equal client allocation}
In this experiment, we compare the performance of FedDAP with other SOTA methods under equal client allocation. Specifically, we allocate $12$ clients for each dataset and distribute an equal number of clients for each domain. As shown in Tables~\ref{comparison_domainnet_office_equal} and~\ref{comparison_pacs_equal}, FedDAP illustrates the significant improvements on all datasets. These results demonstrate the robustness and generalization capability of FedDAP across a domain shift scenario. Overall, the consistent gains under equal allocation highlight the effectiveness of our domain-aware design, which enables better alignment and representation learning across heterogeneous client distributions.

\begin{table}[]
\centering
\resizebox{\columnwidth}{!}{%
\begin{tabular}{l c c c c|c c}
\toprule
\multirow{2}{*}{\textbf{Method}} 
  & \multicolumn{6}{c}{\textbf{PACS}} \\ 
\cmidrule(lr){2-7}
  & Photo & Art & Cartoon & Sketch & Avg & $\Delta$ \\
\midrule
FedAvg    & 86.39 & 82.98 & 93.97 & 90.06 & 88.35 & –    \\
FedProx   & 83.47 & 83.13 & 93.54 & 90.46 & 87.65 & -0.70\\
MOON      & 87.58 & 83.96 & 92.04 & 90.50 & 88.52 & +0.17\\
COPA      & 78.90 & 77.64 & 94.87 & 85.00 & 84.10 & -4.25\\
FedGA     & 86.32 & 83.33 & 95.07 & 89.24 & 88.49 & +0.14\\
FedProto  & 88.28 & 82.95 & 95.35 & 91.26 & 89.46 & +1.11\\
FPL       & 85.85 & 82.28 & 94.37 & 91.86 & 88.59 & +0.24\\
FedPLVM   & 85.84 & 83.25 & 94.88 & 92.75 & 89.18 & +0.83\\
FedRDN
          & 86.19   & 84.72   & 94.64    & 95.17   & 90.18    & +1.83    \\
\midrule
\textbf{FedDAP}
          & 90.53 & 87.18 & 96.57 & 93.33 & \textbf{91.90} & \textbf{+3.55} \\
\bottomrule
\end{tabular}%
}
\caption{Comparison on the PACS dataset under equal client allocation. “Avg” denotes the mean accuracy across all domains, and $\Delta$ indicates improvement over the FedAvg baseline.}
\label{comparison_pacs_equal}
\end{table}

% \begin{table}[H]
% \centering
% \resizebox{\columnwidth}{!}{%
% \begin{tabular}{lcccc|c}
% \toprule
% \multicolumn{6}{c}{\textbf{Digits}} \\
% \textbf{Local Prototypes} & MNIST & USPS & SVHN & SYN & Avg \\
% \midrule
% \texttt{w} DP   & 97.40 & 90.08 & 72.50 & 69.27 & 82.31 \\
% \texttt{w/o} DP & 97.64 & 89.31 & 75.35 & 70.15 & \textbf{83.11} \\
% \midrule
% \multicolumn{6}{c}{\textbf{Office-10}} \\
% \textbf{Local Prototypes} & Caltech & Amazon & Webcam & DSLR & Avg \\
% \midrule
% \texttt{w} DP   & 69.20 & 81.58 & 67.24 & 68.00 & 71.51 \\
% \texttt{w/o} DP & 73.75 & 82.42 & 68.62 & 65.33 & \textbf{72.53} \\
% \midrule
% \multicolumn{6}{c}{\textbf{PACS}} \\
% \textbf{Local Prototypes} & Photo & Art & Cartoon & Sketch & Avg \\
% \midrule
% \texttt{w} DP   & 88.47 & 75.00 & 77.07 & 93.96 & 83.63 \\
% \texttt{w/o} DP & 90.43 & 75.28 & 77.64 & 95.16 & \textbf{84.63} \\
% \bottomrule
% \end{tabular}%
% }
% \caption{Impact of differential privacy on local prototypes across different datasets.}
% \label{prototype_dp}
% \end{table}


\end{document}